\documentclass[10pt,twocolumn,letterpaper]{article}
\usepackage{fancyhdr}
\usepackage{wacv}
\usepackage{times}
\usepackage{epsfig}
\usepackage{graphicx}
\usepackage{amsmath}
\usepackage{amssymb}
\usepackage{float}
\usepackage{booktabs}
\usepackage{multirow}

\wacvfinalcopy 

\def\wacvPaperID{0101} 

\ifwacvfinal\fi
\begin{document}

\title{Video Summarization via Actionness Ranking}

\author{Mohamed Elfeki$^1$ and Ali Borji \\
$^1$University of Central Florida, Center for Research in Computer Vision (CRCV) \\
{\tt\small elfeki@cs.ucf.edu, aliborji@gmail.com} 
}

\maketitle
\ifwacvfinal\thispagestyle{empty}\fi

\fancyhf{}
\cfoot{\thepage}

\lhead{COUCOU}

\begin{abstract}
   To automatically produce a brief yet expressive summary of a long video, an automatic algorithm should start by resembling the human process of summary generation. Prior work proposed supervised and unsupervised algorithms to train models for learning the underlying behavior of humans by increasing modeling complexity or craft-designing better heuristics to simulate human summary generation process. In this work, we take a different approach by analyzing a major cue that humans exploit for summary generation; the nature and intensity of actions. 
   
   We empirically observed that a frame is more likely to be included in human-generated summaries if it contains a substantial amount of \textit{deliberate} motion performed by an \textit{agent}, which is referred to as \emph{actionness}. Therefore, we hypothesize that learning to automatically generate summaries involves an implicit knowledge of actionness estimation and ranking. We validate our hypothesis by running a user study that explores the correlation between human-generated summaries and actionness ranks. We also run a consensus and behavioral analysis between human subjects to ensure reliable and consistent results. The analysis exhibits a considerable degree of agreement among subjects within obtained data and verifying our initial hypothesis. 
   
   Based on the study findings, we develop a method to incorporate actionness data to explicitly regulate a learning algorithm that is trained for summary generation. We assess the performance of our approach on 4 summarization benchmark datasets, and demonstrate an evident advantage compared to state-of-the-art summarization methods.\footnote{Accepted as an oral presentation in WACV-19.}
\end{abstract}



\section{Introduction}
With the immense growth in the use of smart-phones and cameras, the amount of recorded visual data has become by far much more available than what can be attentively viewed. Each day 144,000 hours of video are uploaded to YouTube, which is almost 17 years worth of videos~\cite{intro_1,intro_2,seqDPP}. Moreover, recent statistics report that 245 million CCTV cameras are professionally installed around the world, actively surveying day-to-day activities~\cite{hirsch2017seizing}. Records in 2017 show that there are at least 2.32 billion active camera phones~\cite{obile2016ericsson}. Estimates show that about 2.4 million GoPro body cameras were sold world-wide in 2015~\cite{bizjournals}. This calls for efficient and automatic methods that quickly examine visual data and provide an informative briefing about the original videos. Video summarization addresses the problem of selecting a subset of video frames such that summary captures the most important and representative events of the original video.

\begin{center}
\begin{figure}
\includegraphics[width=0.45\textwidth,height=0.15\textheight]{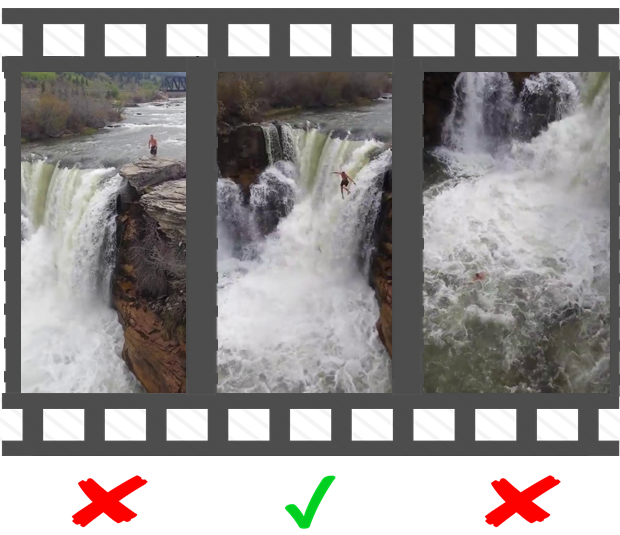}
\caption{When generating summaries, humans often favor frames containing deliberate motion (such as a jumping man) over frames without \textbf{deliberate} motion (such as waterfall), even when natural/non-deliberate motion is more intense. The main question addressed here is whether we can gain insights from learning to recognize deliberate actions (i.e., actionness) to further assist video summarization.}
\label{fig:problem_1}
\end{figure}
\end{center}

\begin{figure*}
\includegraphics[width=\textwidth,height=0.3\textheight]{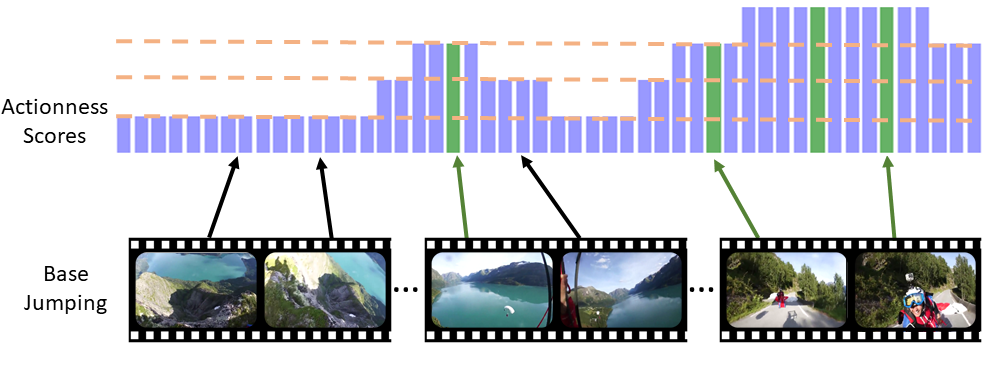}
\vspace{0.11em}
\caption{When examining human-generated summaries, we observe that they usually contain high degree of deliberate actions. In this work we put forth and examine the following hypothesis: \textit{"Frames containing high magnitude of \textbf{deliberate} motion have a higher likelihood of being included within the video summary"}.}
\label{fig:problem}
\end{figure*}

Several prior works made substantial efforts to better understand the video summarization problem and have proposed heuristic solutions (e.g., \cite{discovering_people,story_driven,information_sum,event_sum,attention_sum,category-specific}). The remarkable success of deep neural networks \cite{alexnet,inception,rnn_language,rnn_speech} has motivated researchers in designing even more complex black-box models instead of a developing a profound understanding of the problem (e.g.,~\cite{gan_summarization,zhang_lstm,bmvc_summ_caption,encoder_decoder}). While increasing model complexity often helps in better modeling the latent patterns of data, it has the risk of overfitting to standard benchmark training video datasets and being sensitive to noise and irrelevant features, unless a proper learning objective is used. To address this challenge, here we seek to investigate a new learning objective that takes into account the role of deliberate actions performed by generic agents within the human-generated summaries and utilize this correlation to perform a robust automatic summarization. The premise of our work stems from our observation that humans tend to include frames with deliberate actions more frequently in the summary, since they tend to represent more ``\textit{unexpected and important}" events, and tell more about the story of the video.

Actions and motion patterns in videos present an intricate visual stimulation to the eyes of the viewer and thus become major cues when generating summaries for long videos. In the philosophy of actions~\cite{actions_1963}, there are three aspects that define a generic action instance: i) it is carried out by an \textit{agent}, ii) it requires an \textit{intention}, and iii) it leads to \textit{side-effects}. \textit{Spatial Actionness} was introduced to quantify the likelihood of an image region to contain a generic action instance ~\cite{actionness_ranking,actionness_2016}. Along the same lines, video summarization aims to localize temporal instances where important events occur. We propose to extend this definition to the temporal domain to better serve the summarization problem. That is, \textit{Temporal Actionness} is the likelihood of a generic action to appear within a temporal video segment. 

Temporal actionness ranking can assist an automatic summarization algorithm in localizing and quantifying the intensity of generic action instances. Consequently, it can also estimate the likelihood of including each event in the summary. Fig. \ref{fig:problem} shows an example of a first-person video of a person performing base jumping. There are four distinct types of motion in this video: \textit{running water, camera relative motion, a jumping partner, and first-person own-hand manipulation}; but only the last two instances qualify as strong temporal actionness which tend to constitute the vast majority of the summary.

Our main contributions in this paper is three-fold. First, we establish the concept of temporal actionness and study how it relates to video summarization. Second, we introduce a new set of actionness labels over four existing summarization benchmarks, and run a consensus and behavioral analysis on them to verify their consistency. Finally, we propose a method that utilizes temporal actionness to improve the summary generation through a multi-task learning formulation.

\section{Related Work}
In this section, we start by reviewing the concept of spatial actionness in the literature. Then, we briefly review Recurrent Neural Networks (RNN) and mention some of their applications in video processing. Finally, we conclude by discussing some prior approaches that have applied RNN models to the video summarization problem.

\noindent\textbf{Actionness:} The concept of spatial actionness was first introduced in \cite{actionness_ranking} as the deliberate bodily movement performed by an agent; which is distinct from general instances of motion since it requires intention. They used Lattice Conditional Ordinal Random Fields to rank the regions of an image based on its likelihood of containing an action (i.e., ranking actionness).

Accurate and efficient ranking of spatial actionness was shown to benefit other related tasks~\cite{actionness_2016,actionness_localization,actionness_recognition,actionness_proposals}. For example, Wang et al. \cite{actionness_2016} used a fully convolutional network to estimate spatial actionness. Then, they embedded the predicted actionness heat-map within a hybrid approach that performs action detection. Also, Ting et al. \cite{actionness_localization,rebuttal_2} suggested a framework that performs action proposals by generating actionness curves via a snippet-level actionness classifier, then grouping them over time to produce the proposal candidates. Finally, Zhao et al. \cite{rebuttal_1} proposed a temporal action proposal scheme called Temporal Actionness Tagging (TAG). This method uses an actionness classifier to evaluate the binary actionness probabilities for individual snippets. Our definition of temporal actionness is consistent with theirs, but also generalizes to agents other than humans as discussed in Section 3.1.

\noindent\textbf{Recurrent Neural Networks (RNNs):} Since their introduction in \cite{rnn,bptt}, RNNs have been commonly used to model sequential data. Unlike feed-forward networks (e.g., CNNs) whose output only depends on the input at the current time-step, RNN output also relies on previous time-steps. The basic formulation of RNN has the drawback of missing long-term dependencies due to the vanishing gradient problem \cite{vanishing_gradient}. Several extensions of RNNs have been introduced to resolve this problem. Popular approaches include: Long-Short Term Memory (LSTM) \cite{lstm}, and Gated Recurrent Unit (GRU) \cite{gru}. Both of these models have been successfully employed for applications such as video captioning using LSTM~\cite{cap_1,cap_2,cap_3,cap_4}, and action recognition and action proposals using GRU~\cite{sst,sst_AD,act_gru}. 

\noindent\textbf{Video Summarization using RNNs:} Because of their ability to process temporal data, RNNs have been widely used to train supervised and unsupervised video summarization models (e.g., ~\cite{encoder_decoder,gan_summarization,zhang_lstm,bmvc_summ_caption,lstm_1,lstm_2}). Zhang, et al.~\cite{zhang_lstm} were the first to use a supervised LSTM and a Multi-Layer Perceptron (MLP) while optimizing the Determinantal Point Process (DPP) maximum likelihood~\cite{dpp_1,dpp_2,dpp_3,seqDPP}. DPP is used to quantify the diversity in the selected subset of frames which deems maximizing DPP to be equivalent to selecting a representative summary since the redundancy is minimized. Recently, Mahesseni et al.~\cite{gan_summarization} presented an unsupervised video summarization framework by training an LSTM network in an adversarial manner to better model the complexity of the data. 
Further, Chen et al.~\cite{bmvc_summ_caption} used a hybrid framework that utilizes GRU, MLP, and a temporal segmentation algorithm to perform the tasks of video summarization and video captioning simultaneously.

\section{Relating Actionness to Summarization}
In this work we hypothesize that human-generated summaries favor frames that contain deliberate motions over stationary or monotonous motions that are deemed boring. To test this hypothesis, we start by defining the type of motion that we expect to be a substantial component in human-generated summaries, which we refer to as temporal actionness. Then, we conduct a user study on human subjects investigating the relationship between temporal actionness and generated summaries. Finally, we conduct a consensus analysis on the obtained data to measure the agreement among subjects and a behavioral analysis to ensure the reliability of our findings.

\subsection{Temporal Actionness}
As discussed in Section 2.1, spatial actionness is defined as the likelihood of a certain region in an image to contain an action~\cite{actionness_ranking}. An image region is considered to contain an action based on the definition of actions in~\cite{actions_1963} as "what an agent can do with a deliberate bodily movement that leads to side-effects".

Our definition of actionness is consistent with the aforementioned definitions, but we extend it in two ways. First, we also consider non-human agents that perform deliberate motions, because human agents do not necessarily exist in the videos that are required to be summarized. For example, a swimming dolphin represents an action while a running river is not. Even though both of them contain similar magnitudes of motion but there is no intention in the latter.

Second, we adapt the actionness concept to the temporal domain, where we estimate the likelihood of a given video segment to contain an action. For biological agents, it is possible to predict the likelihood of the action from the agent's pose. However, since we are generalizing our definition to non-biological agents, their motion often is not distinguishable within a single frame. Thus, a video segment is essential to determine the nature of motion. For instance, detecting a moving vehicle requires monitoring several frames to track the vehicle's location changes and to distinguish it from a stopped one.


We target a rank ordering of actionness rather than a binary classification of whether a segment contains an action (i.e., action proposal~\cite{sst}) for two reasons. First, the fundamental notion of temporal actionness as "localizing when there is an action" immediately presents a difficulty: temporal segmentation remains a challenging and open problem. Some efficient methods exist for this purpose such as KTS~\cite{category-specific}, but the average f-score remains too low for robust use (about 0.41). Ranking makes it more plausible to provide a stratified quantification to the likelihood of a segment based on the prevalence of an action. Second, in any given video, often background actions (e.g., monotonous actions) are overlooked by the viewers as opposed to foreground abrupt actions. For instance, in a surveillance video, it is only natural to dismiss the background monotonous moving traffic, and monitor the abrupt motions around a building's entrance.

\subsection{User Study}
To estimate actionness, we first used KTS algorithm~\cite{category-specific} to produce semantically consistent variable-size segments that contain atomic semantic meanings. Then, for each segment, we asked five users to label it by selecting the appropriate rank from the following scales: \\
0: No action (No deliberate motion by an agent)\\
1: Background action (Weak indication of an action)\\
2: Partial foreground action (Strong action indication covering a minor part of the segment)\\
3: Active foreground action (Strong action indication covering a major part of the segment)\\

For a tractable annotation process, we subsampled the videos to 1 fps. Then, we constructed the displayed segment to contain all the frames in a grid display allowing the users to see all the frames of one segment simultaneously. Before starting the process, users underwent a training stage to understand the task and the procedure. They were asked to rank actionness on four videos. After training, the users were asked to perform the same task on four benchmark summarization datasets: SumMe~\cite{summe}, TVSum~\cite{tvsum}, Youtube~\cite{youtube}, and OVP~\cite{ovp_1}. Videos used during the user training stage were discarded in model development. 

\begin{figure}
\includegraphics[width=0.48\textwidth,height=0.23\textheight]{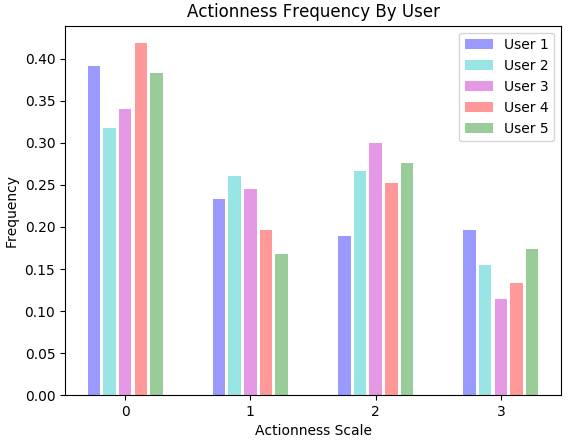}
\caption{\textit{How often each user chose a given actionness scale in the annotations?} Having close frequencies indicates a general agreement between the users.}
\label{fig:annotations_frequency}
\end{figure}

\subsection{Data Analysis}
\noindent {\bf Consensus analysis.}
To ensure the validity of the annotations, we measured the consensus among users using two metrics. The first metric is the f-1 score. We computed the average pairwise f1-measure to estimate the agreement among the annotators for each scale. We obtained 0.55, 0.40, 0.48, and 0.51 for SumMe, TVSum, OVP, and Youtube datasets, respectively. The second metric is the rank-frequency over original videos for each user. That is, how often each user chose a given scale for all the videos of the annotation? Fig. \ref{fig:annotations_frequency} shows the frequency ranks for all users. We observe that ratios by users are close to each other for all the scales, which along with the f-1 scores demonstrates evident consensus among users.

\begin{figure}
\includegraphics[width=0.48\textwidth,height=0.23\textheight]{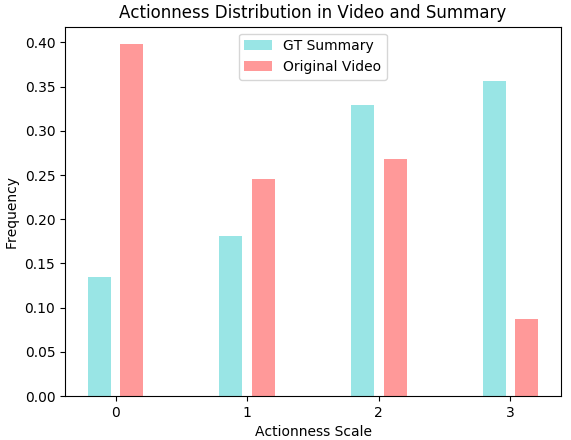}
\caption{\textit{Do GT summaries contain high actionness?} GT summaries mostly consist of scale-three actionness, while original videos mostly contain scale-zero actionness.}
\label{fig:actionness_ratio}
\end{figure}

\noindent {\bf Do summaries contain high actionness?}
To answer this question, we computed the average frequency of each actionness scale in both of the ground-truth summary and the original video. Fig. \ref{fig:actionness_ratio} demonstrates that scale-three actionness frames seem to be the dominant majority rank among the summary despite their minority existence in the original video. \emph{Hence, frames containing high actionness are more likely to be included in the summary}.

\begin{figure}
\begin{center}
\includegraphics[width=0.48\textwidth,height=0.23\textheight]{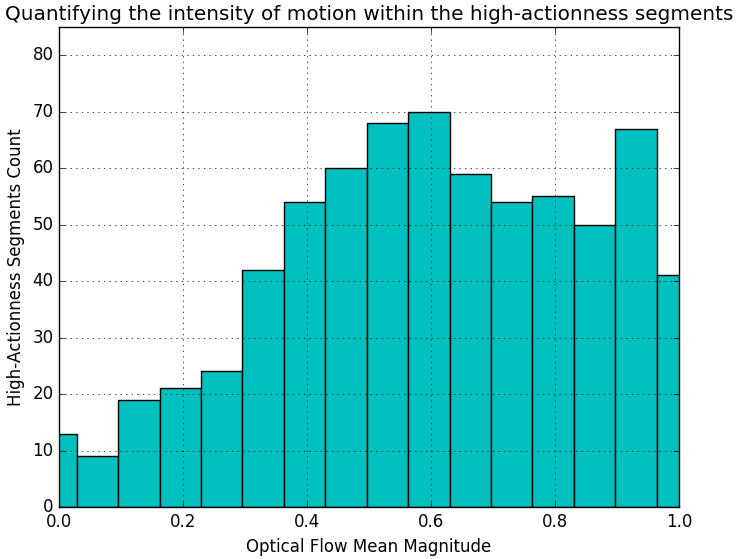}
\vspace{0.1em}
\caption{\textit{Were the annotators just looking for abrupt motions?} Non-abrupt motions also exists vastly in the selected summaries, mostly with optical flow changes $\geq$ 25\%}    
\end{center}
\label{fig:deliberate_experiment}
\end{figure}

\noindent {\bf Were the annotators just looking for abrupt motions?}
For a more extensive verification, we examine if the users tended to choose segments containing abrupt motion (i.e., high magnitudes of motion) as representation for the high-actionness segments. To answer this question, we first need to provide an evaluation for abrupt motion. We calculated the mean magnitude of optical flow for each of the segments, and normalized it across each video. Then, we computed the histogram plot of the segments scored by the users as level-three actionness sorted by their normalized mean magnitude of optical flow. As shown in Fig.~\ref{fig:deliberate_experiment}, the selected segments are distributed among a wide variation of optical-flow intensities. This shows that users were not merely selecting the most abrupt motion segments as representatives for the deliberate actions required in high actionness.

\begin{figure*}[!ht]
\centering
\includegraphics[width=0.9\textwidth,height=0.3\textheight]{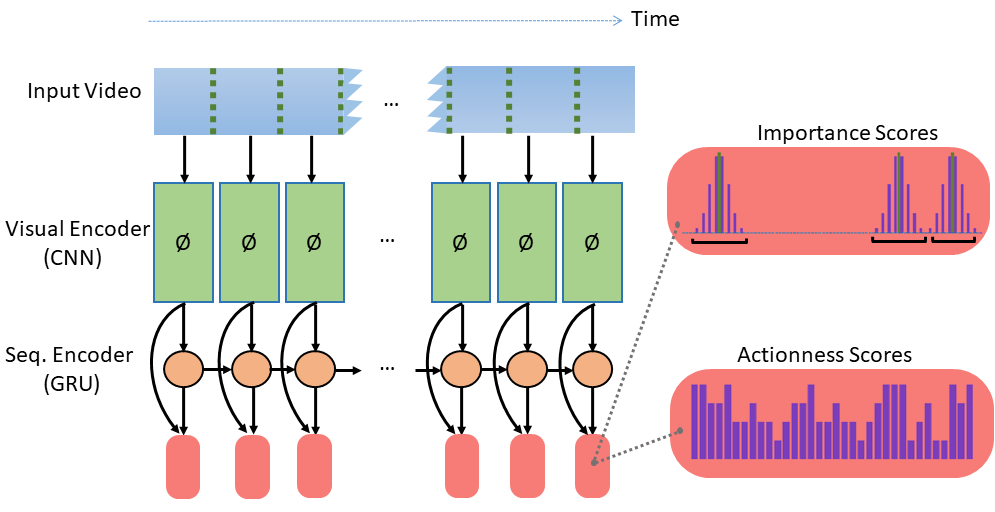}
\vspace{1.67em}
\caption{Using actionness ranking (i.e., actionness level classification of each frame) to regularize the learning of video summarization. Summarization is learned by maximizing diversity within the selected subset. Training the framework in a multi-task learning fashion with an accessory task of actionness ranking, improves the learning of the main task (i.e., video summarization).}
\label{fig:method}
\end{figure*}

\noindent {\bf Oracle labels.}
Having established our hypothesis, we seek to utilize the data obtained from the study to further improve the automatic video summarization algorithms. In order to train a supervised learning model, we need to produce a single set of labels out of multiple annotations for each video. This is often referred to as \textit{Oracle Labels} set. We follow the algorithm proposed in~\cite{seqDPP,dpp_3} that greedily selects the segment that results in the largest marginal gain on the f-1 score computed between the users' annotations. To produce frame-level labels, we consider all the frames within a segment to have its ranking label.\\

\section{Approach}
In this section we propose a model that incorporates actionness ranking task to regularize video summarization.

\subsection{Overview}
Figure \ref{fig:method} shows an overview of our framework. The input is a video of $n$ frames. First, a visual encoder $\phi$ (i.e., a pretrained CNN) is used to extract spatial features for each frame. Next, the extracted features are sent to a sequential encoder (i.e., a Bi-directional GRU) to extract their corresponding temporal features. GRU is used as a sequential encoder because it has fewer parameters than LSTM, which results in faster training and a less risk of overfitting, and shown to perform on par to the LSTM~\cite{gru}. Next, we aggregate both types of the features, spatial and temporal, to generate a comprehensive spatio-temporal feature vector for each frame. These features represent the visual information of the current frame as well as encode all the temporal information from other frames in the video. Finally, the aggregate features are mapped to the actionness and importance scores using two independent MLPs. 

The framework is trained to learn two tasks: 1) summarization by minimizing importance estimation loss, and 2) actionness ranking by minimizing actionness classification loss. The framework is optimized by applying a regularized multi-task learning paradigm~\cite{regularized_multi_task}. Imposing a regularization term in a joint loss is aimed to penalize the unnecessary complexity of the original learning problem that might cause overfitting to training data, while enforcing learning task relationship.

By combining the two losses into a single joint loss, the network is trained to learn a set of trainable parameters $\theta$ such that:
\begin{equation}
argmin_{\theta} \ \  S(\theta) + \lambda R(\theta),
\label{eq:joint_loss}
\end{equation}
where $S(\theta)$ is the summarization loss (section 4.2), $R(\theta)$ is the actionness classification loss (section 4.3) which acts as a regularizer, and $\lambda$ is the regularization weight used to force both the losses to operate on comparable ranges, preventing the learning to be biased towards one of the losses.

\subsection{Importance Estimation}
Importance scores (i.e., summarization labels) are binary labels that indicate the frames selected to be a part of the summary: 1 for selected frames, and 0 otherwise. The problem with this type of labeling is that frames within the same segment tend to have similar semantic features, therefore the annotators could have chosen any other frame within a selected frame's segment (i.e., key segment). To reduce the effect of the inherent noise in the labels, we apply Gaussian smoothing as a preprocessing step. Particularly, binary labels are converted to real-values where the mean is the selected frame within the summary, and the Gaussian distribution is sampled across its key segment (see Fig. \ref{fig:method}). Thus, the framework would not be penalized for choosing a frame within a key segment as much as it would be penalized for choosing a frame outside a key segment.
 
Increasing the diversity within the selected subset is equivalent to choosing a representative subset since the redundancy is minimal. Following \cite{seqDPP,dpp_2}, we follow the decomposition in~\cite{dpp_3} to compute the marginal kernel $L_y$ as a of a Gram matrix in the following manner:
\begin{equation}
L_{ij}= q_i\phi_i^\top\phi_j q_j;\,\, \forall i, j\in y
\label{eq:dpp_decomposition}
\end{equation}

where $\phi_i$ can be seen as a representative feature vector, and $q_i$ is quality score of frame $i$ in the selected subset $y$. Similar to~\cite{zhang_lstm}, we construct the features with a dimensionality of 256 for each frame, and the quality score as a single scalar for every frame. In our framework, we apply two independent MLPs with the aforementioned dimensions to obtain $\phi$ and $q$ and compute the marginal DPP kerenl as in Eq.~\ref{eq:dpp_decomposition}.

Finally, we optimize the Maximum Likelihood Estimation (MLE) of the normalized marginal DPP kernel that quantifies the diversity in the ground-truth summaries $y$ as follows:
\begin{equation}
S(\theta)= -log\left[\frac{det(L_y)}{det(L+I)}\right]
\end{equation}
where $L$ is the marginal kernel of the ground-set of all the frames in the video, and $I$ is the identity matrix.

\subsection{Actionness Ranking}
This task aims to provide a regularization term to the joint loss (Eq. \ref{eq:joint_loss}) which is determined by classifying the actionness scale $v$ of each frame; $v \in \{0, 1, 2, 3\}$. We train an independent MLP to map the spatio-temporal features of each frame to an actionness rank using the categorical cross entropy loss as follows:
\begin{equation}
R(\theta) = - \sum_{i=1}^n \sum_{j=0}^{3} t_{i,j} log(p_{i,j}),
\label{eq:joint_loss_2}
\end{equation}
where $p_{i,j}, t_{i,j}$ are the predicted and target values of actionness rank $j$ for the $i$-th frame.

\section{Experimental Results}
In above sections, we proposed that deliberate motion provides a significant cue when humans are summarizing a given video. Then, we established this hypothesis by performing a user study among multiple human subjects that were asked to rank the magnitude of deliberate motion. By analyzing the study results, it is clear that a significant portion of the summary includes high intensity of deliberate motion, as opposed to the original video contents. Therefore, we introduced an approach that can rank the intensity of deliberate motion and uses this knowledge to improve the performance to perform a better video summarization. In this section, we run an extensive set of experiments where we show the effect of learning the actionness in learning summarization.

\subsection{Datasets}
We evaluated our approach on four summarization benchmark datasets: SumMe~\cite{summe}, TVSum~\cite{tvsum}, Open Video Project (OVP)~\cite{ovp_1}, and Youtube~\cite{youtube}. The first dataset consists of 25 user videos covering multiple events such as bears climbing a tree and cooking. It contains both first-person and third-person videos with lengths varying from 1.5 to 6.5 minutes. The second dataset consists of 50 Youtube videos from 10 categories of the TRECVid Multimedia Event Detection (MED), 5 videos per category. They vary in length from 1 to 5 minutes and include both first and third person videos.

The third and fourth datasets are quite large. We use the same subset of videos used in~\cite{youtube,gan_summarization,zhang_lstm}, 50 videos from OVP, and 39 videos from Youtube. OVP videos contain mostly news reports and documentary clips that vary in length from 1 to 4 minutes. All of them are third-person videos. The last dataset contains news and sports videos (third-person videos) with lengths varying from 1 to 10 minutes.

\begin{table*}[]
\centering
\resizebox{0.8\textwidth}{3.2cm}{%
\begin{tabular}{@{}ccccccccc@{}}
\toprule
\multirow{2}{*}{Model} & \multicolumn{2}{c}{Canonical} &  & \multicolumn{2}{c}{Augmented} &  & \multicolumn{2}{c}{Transfer} \\ \cmidrule(lr){2-3} \cmidrule(lr){5-6} \cmidrule(l){8-9} 
 & SumMe & TVSum &  & SumMe & TVSum &  & SumMe & TVSum \\ \midrule
{[}29{]} & 26.6 & - &  & - & - &  & - & - \\
{[}15{]} & 39.7 & - &  & 39.7 & - &  & - & - \\
{[}14{]} & 39.5 & - &  & 39.3 & - &  & - & - \\
{[}55{]} & 40.9 & - &  & 40.9 & - &  & 38.5 & - \\
{[}56{]}-vsLSTM & 37.6 & 54.2 &  & 37.6 & 54.2 &  & 41.6 & 57.9 \\
{[}56{]}-dppLSTM & 38.6 & 54.7 &  & 38.6 & 54.7 &  & 42.9 & 59.6 \\
{[}33{]}-DPP & - & - &  & 39.1 & 51.7 &  & 43.4 & 59.5 \\
{[}33{]}-Sup & - & - &  & 41.7 & 56.3 &  & 43.6 & \textbf{61.2} \\ \midrule
Ours-Basic & 37.9 & 54.6 &  & 38.8 & 54.8 &  & 43.1 & 59.6 \\
Ours-FT & 38.7 & 54.9 &  & 42.3 & 56.1 &  & 43.8 & 59.3 \\
\textbf{Ours-Reg} & \textbf{40.1} & \textbf{56.3} & \textbf{} & \textbf{45.8} & \textbf{59.1} & \textbf{} & \textbf{46.1} & 60.1 \\ \bottomrule
\end{tabular}%
}
\vspace{1.67em}
\caption{F1-scores for several test configurations. Canonical: Train on 80\% of a dataset, test on the remaining 20\%. Augmented: Train on one dataset, test on the other. Transfer: Train on one dataset + OVP + YouTube, test on the other.}
\label{tab:system_performance}
\end{table*}

\subsection{Experimental Setup}
For a fair comparison with the related approaches, we evaluate our method using the keyshot-based metric similar to~\cite{zhang_lstm,gan_summarization}. We first convert frame-level scores to shot scores by applying the KTS algorithm~\cite{category-specific} that generates semantic shots. The resulting shots are ranked based on their importance score, which is the average score of the frames in that shot. By applying the Knapsack algorithm, a subset of the highest ranked keyshots are selected such that the total duration of the generated summary is less than 15\% of the original video. We report the average f1-scores to evaluate the predicted summary as compared to the ground-truth summary.

\noindent \textbf{Implementation Details:} Similar to~\cite{gan_summarization,zhang_lstm}, we use the output of the pool5 layer of GoogLeNet~\cite{googlenet} architecture trained on ImageNet~\cite{imagenet} as the visual encoder for our framework to extract a 1024 dimension spatial feature vector for each frame. Then, we use a single-layer GRU with 256 hidden units as the sequential encoder and 256 hidden units MLPs for both of the optimization tasks. Similar to the training setup of~\cite{zhang_lstm}, we run our model for 100 iterations in the training stage and stop the training if the validation f1-score does not improve for more than 5 consecutive iterations. The validation split is set to be 20\% random subset of the training data. We use Adam optimizer to train our framework with learning rate of 0.001. To learn the task of actionness ranking, we set $\lambda$ to 0.003. The value of $\lambda$ was selected to make both of the losses operate on close ranges so that none of them bias the optimization while training the network.

\subsection{System Performance}
\noindent \textbf{Test Configurations:} We follow \cite{zhang_lstm,gan_summarization} to evaluate our method in three test configurations. In the first configuration (Canonical), we use 80\% of one dataset to train the method, and test the method on the remaining 20\% of the same dataset.
In the second configuration (Augmented), TVSum and SumMe datasets are used together - one dataset is used to train the method while being tested on the entire other dataset. 
In the last configuration (Transfer), we adapt the same paradigm as the second configuration but augment the training set with OVP and Youtube datasets, which improves the results on SumMe and TVSum. 

\noindent \textbf{Baselines:} We conduct an extensive comparison with the state of the art methods \cite{summe,mixture,summary_transfer}, two models from~\cite{zhang_lstm}: LSTM+MLP (vsLSTM) and LSTM+MLP+DPP (dppLSTM), and two models from ~\cite{gan_summarization}: Unsupervised DPP (DPP) and supervised model (SUP).

Also, to perform an ablation study on our model, we introduce three variants of our approach. First, \textit{Ours-Basic} is our model without the actionness regularization;. It reduces the model's complexity to be close to~\cite{zhang_lstm}, however, our model uses GRU instead of LSTM and performs Gaussian smoothing preprocessing on the labels. Second, \textit{Ours-FT} is the same as the basic model, but the sequential encoder is first trained for human-based action localization, then the entire framework is fine-tuned for video summarization. To train the GRU for action localization, we follow~\cite{rnn_localization} to train the sequential encoder on GoogLeNet features for action recognition task on UCF-101 ~\cite{ucf101} for 100 epochs, then fine-tune it for action localization on THUMOS-14~\cite{thumos14} for another 100 epochs. The last model is \textit{Ours-Reg}, which is a model that is trained for simultaneous video summarization and actionness estimation as discussed in Section 4.


\noindent \textbf{Summarization Evaluation:} Table~\ref{tab:system_performance} shows the f-1 scores of our models compared to the state-of-the-art methods. As shown, Ours-Basic performs similarly to vsLSTM and dppLSTM. Training our model on the action recognition labels prior to summarization (Ours-FT) performs on par with the state-of-the-art methods. However, the model that is trained for actionness estimation, that is considering deliberate motions performed by generic agents (not just humans unlike Ours-FT), significantly outperforms all other methods in most of the settings (Ours-Reg).

\begin{figure}[t]
{
  \centering
   \includegraphics[width=.5\textwidth,height=0.3\textheight]{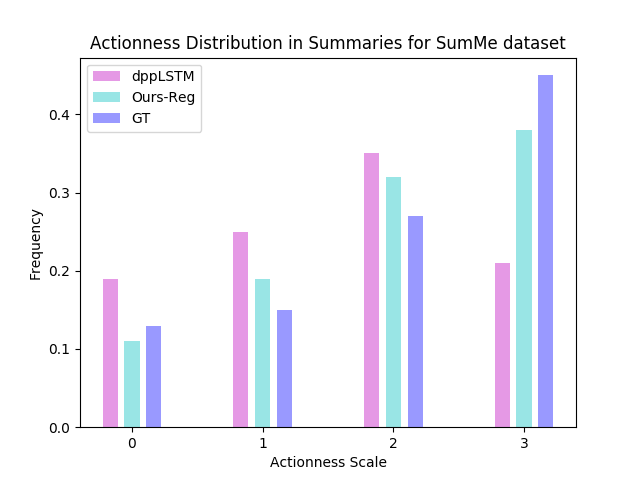}

   \label{fig:qual}
   \caption{Distribution of actionness scales over summaries of SumMe dataset. Our model better resembles the GT than dpp-LSTM~\cite{zhang_lstm}.}
}
\end{figure}

\noindent \textbf{Actionness Evaluation:} To investigate whether actionness helps summarization, we ran two analyses. First, we verify that our model effectively learns the actionness ranking task by computing the actionness classification accuracy in all test configurations. As shown in Table ~\ref{tab:classification}), Ours-Reg performs significantly better than chance, indicating that the model actually learns actionness estimation and does not dismiss it from the learning procedure. Second, we compute the distribution of actionness scales in the ground-truth summary, Ours-Reg, and~\cite{zhang_lstm} over the SumMe dataset for test configuration 1 (see Fig. 5). As shown in Fig.~\ref{fig:qual}, our model resembles the ground-truth summary better than ~\cite{zhang_lstm}. The two results suggest that learning actionness ranking is indeed useful for better video summarization.

\begin{table}[t]
\centering
\begin{tabular}{@{}ccc@{}}
\toprule
 & SumMe & TVSum \\ \midrule
\textit{Chance} & \textit{36.6} & \textit{28.1} \\ \midrule
Canonical & 39.7 & 30.3 \\
Augmented & 42.8 & 32.6 \\
Transfer & 41.8 & 29.5 \\ \bottomrule
\end{tabular}
\vspace{1.67em}
  \caption{Actionness Classification Accuracy of Ours-Reg: In all the settings our model learned to estimate actionness better than the chance level.}
  \label{tab:classification}
\end{table}

\section{Conclusion and Future work}
In this work, we present a further step in analyzing and understanding the video summarization problem. We hypothesize that humans actively rely on \textbf{deliberate} motion and action cues -among other cues- to generate a brief summary that best expresses long visual sequences. We examine this hypothesis by running a user study, investigating the correlation between human-generated summaries and actionness ranking. We then conduct a consensus and behavioral analysis on the data obtained from users to ensure the data reliability and agreement among the users. The findings of the study show a substantial likelihood of including frames containing high actionness ranks within the summaries. 

Thus, we propose a new method that utilizes actionness cues to better learn the task of video summarization. We use a recurrent neural network that is trained for video summarization while being explicitly regularized to learn the actionness ranking task in a multi-task learning formulation. The evaluation on four benchmark summarization datasets shows a significant improvement by our approach over several state-of-the-art summarization methods.

\textbf{Future Work:} The main objective of this work was to examine the relationship between the tasks of actionness estimation and video summarization, and using the former to improve the performance of the latter. As the initial step, we used an extra set of annotations called actionness to train a summarization model in a supervised manner. For future work, we plan to utilize the actionness information to train simpler more efficient video summarization methods in an unsupervised manner.

\noindent \textbf{Acknowledgment.} We would like to thank Professor Abhijit Mahalanobis for helpful discussions and feedback, and NVIDIA for donating the GPU used in the experiments.

{\small
\bibliographystyle{ieee}
\bibliography{egbib}
}

\end{document}